\documentclass[cameraready]{Interspeech}
\title{Benchmarking Automatic Speech Recognition Tools for Iberian Languages}

\author[affiliation={1, 2}, orcid=0000-0002-7705-2250, correspondingauthor]{Fernando}{López}
\author[affiliation={1}]{Pablo}{Gómez}
\author[affiliation={1}, orcid=0000-0001-6979-9330]{David}{Solans}
\author[affiliation={1}, orcid=0000-0001-9293-8052]{Paulo}{Villegas}
\author[affiliation={1}, orcid=0000-0002-4507-4930,]{Jordi}{Luque}

\address{
    $^1$ Telefónica Innovación Digital, Spain \\
    $^2$ Universidad Autónoma de Madrid, Spain
}

\email{fernando.lopez@telefonica.com}


\usepackage{comment}

\begin{document}

\maketitle

\begin{abstract}
    Comprehensive evaluations of automatic speech recognition (ASR) for Iberian languages remain limited, and low-resource languages, biases, and efficiency trade-offs are underexplored. We benchmark eleven systems, ten open-weight models and one commercial API, across five Iberian languages (Basque, Catalan, Galician, Portuguese, Spanish), with German and Turkish as controls. Evaluation uses an 85-hour dataset covering read speech, broadcast media, and audiobooks, assessing accuracy and efficiency via word error rate (WER) and real-time factors (RTF/RTFx). Results show no single model dominates: accuracy, efficiency, and language coverage present clear trade-offs. Low-resource languages, especially Basque, degrade significantly, highlighting the role of training coverage. We observe consistent sex disparities across most systems, highlighting fairness challenges in multilingual ASR. Overall, the benchmark provides practical guidance for real-world model selection.
\end{abstract}

\section{Introduction}
Automatic speech recognition (ASR) has advanced rapidly. Public datasets \cite{panayotov2015librispeech, pratap2020mls, omnilingual2025omnilingual} and open pre-trained models \cite{radford2023robust, baevski2020wav2vec, hsu2021hubert, omnilingual2025omnilingual} have expanded quickly, accelerating both development and deployment. Large-scale ASR evaluations, however, remain heavily skewed toward English: even benchmarks with a multilingual track tend to prioritize short-form English and cover only a handful of high-resource languages \cite{srivastav2025open, conneau2023fleurs}. Consequently, the languages of the Iberian Peninsula receive limited coverage. Although these languages are spoken by millions \cite{baucells2025iberobench}, their support in ASR is uneven. Spanish and Portuguese are comparatively well-resourced, Catalan occupies an intermediate position with growing representation in speech datasets, and Basque and Galician remain notably under-resourced \cite{de2025whisper}.

Several efforts partially address this imbalance. The Albayzin evaluation, held at IberSPEECH, benchmarks speech-to-text transcription for Spanish broadcast media \cite{lleida2022overview} and for bilingual Basque-Spanish code-switched speech \cite{penagarikano24bilingual}, but it does not span the full set of Iberian Peninsula languages. IberoBench \cite{baucells2025iberobench} targets all of these languages, yet it is a multi-task text benchmark for evaluating the natural language understanding capabilities of large language models (LLMs) rather than a speech benchmark. In other geographical regions, dialect-level studies examine single languages in depth, including Arabic \cite{droua2012speaker}, French \cite{maison2023cereales}, and Chinese \cite{wang2026wenetspeech}, underscoring the relevance of fine-grained regional ASR benchmarking.

In short, for the Iberian Peninsula languages, existing work either compares many models across a few languages or covers many languages with a text language understanding focus. 
%
%
Additionally, a solid comparison must also reflect how these systems behave in practice: performance is condition-dependent, so no single model is best everywhere, and results drawn from a single corpus cannot support robust conclusions. 
Word error rate (WER) alone likewise offers an incomplete view and should be paired with metrics that capture computational cost. Aggregate WER can further mask demographic disparities, since ASR systems exhibit measurable bias across speaker groups, with higher error rates repeatedly reported for female speakers and other under-represented populations~\cite{shah2025speech, garg2018word, feng2021quantifying, martin2023bias}. Reporting per-language and per-sex performance is therefore essential to a fair and complete assessment.
%

To close this gap, we make the following contributions. (i) We systematically evaluate diverse ASR systems on the most spoken languages of the Iberian Peninsula, namely Basque, Catalan, Galician, Portuguese, and Spanish, with German and Turkish as contrastive controls\footnote{Data compilation code: \href{https://github.com/ferugit/iberian-asr-bench}{github.com/ferugit/iberian-asr-bench}}; in total, we evaluate ten open-weight systems and one commercial model accessed through an API, reporting WER, real-time factor (RTF), and its inverse (RTFx) to jointly assess accuracy and efficiency. (ii) We analyze the trade-off between recognition accuracy and computational efficiency. (iii) We study per-language and per-sex performance across models, revealing disparities in ASR robustness along linguistic and demographic lines.

\section{ASR benchmark}
We assembled the benchmark in three stages. First, we selected existing datasets for the target languages, preserving their original metadata. Next, we selected a representative set of open-weight models that support either the full range of these languages or specific subsets of them. Finally, for each model we generated transcription hypotheses to assess recognition accuracy and measured inference latency to assess efficiency.

\subsection{Datasets}

\begin{table*}[htbp]
\caption{Evaluated speech datasets with licensing, transcription, and metadata characteristics. ``Sex labels'' indicate whether the source provides speaker sex annotations (Partial: available for a subset of samples).}
\vspace{-0.15cm}
\centering
\small
\resizebox{\textwidth}{!}{%
\begin{tabular}{lccccccccc}
\toprule
Dataset & Language & Hours & Samples & \# speakers & License & Domain & Cased & Punctuation & Sex labels \\
\midrule
SLR 69      & Catalan    & 9.42  & 4240  & 36   & CC BY-SA 4.0 & Read speech     & Yes & Yes & Yes \\
SLR 76      & Basque     & 13.86 & 7136  & 29   & CC BY-SA 4.0 & Read speech     & Yes & Yes & Yes \\
SLR 77      & Galician   & 10.32 & 5587  & 34   & CC BY-SA 4.0 & Read speech     & Yes & Yes & Yes \\
SLR 108     & Spanish    & 10    & 2507  & -    & CC BY 4.0    & Media/broadcast & No  & No  & No \\
SLR 108     & Turkish    & 10    & 2513  & -    & CC BY 4.0    & Media/broadcast & No  & No  & No \\
SLR 94      & Portuguese & 3.74  & 871   & 10   & CC BY 4.0    & Audiobooks      & No  & No  & Yes \\
CommonVoice & German    & 28.03 & 16202 & 5054 & CC0          & Read speech     & Yes & Yes & Partial \\
\bottomrule
\end{tabular}
}
\label{tab:datasets}
\end{table*}

Table~\ref{tab:datasets} summarises the seven datasets. They span roughly 4 to 28 hours per language and differ in audio format and metadata: some provide sex or speaker identifiers, others only the transcription.
\textbf{OpenSLR 69}, \textbf{OpenSLR 76}, and \textbf{OpenSLR 77}~\cite{kjartansson2020open} are crowdsourced, volunteer-recorded read speech for Catalan, Basque, and Galician. They are derived from Wikipedia articles and named-entity templates, and include transcripts and sex metadata. \textbf{OpenSLR 108}~\cite{mediaspeech2021} (MediaSpeech) consists of short, manually transcribed YouTube segments with no metadata beyond a sample identifier; we use the Spanish and Turkish subsets. \textbf{OpenSLR 94}~\cite{pratap2020mls} (Multilingual LibriSpeech) is read audiobook speech from LibriVox; we use the Brazilian Portuguese test split. \textbf{German Common Voice}~\cite{ardila2020common} contains read speech from more than 5k speakers reading predefined sentences; we use the test split exclusively.

Because the datasets differ in domain (read speech, broadcast speech, and audiobooks), cross-language differences conflate intrinsic language difficulty with domain and recording conditions, and this design cannot separate the two. Turkish and German serve as controls for differing levels of resource availability, with Turkish under-resourced relative to German.

\subsection{Data statistics}
\begin{figure}[h]

\centering
\includegraphics[width=1.0\linewidth]{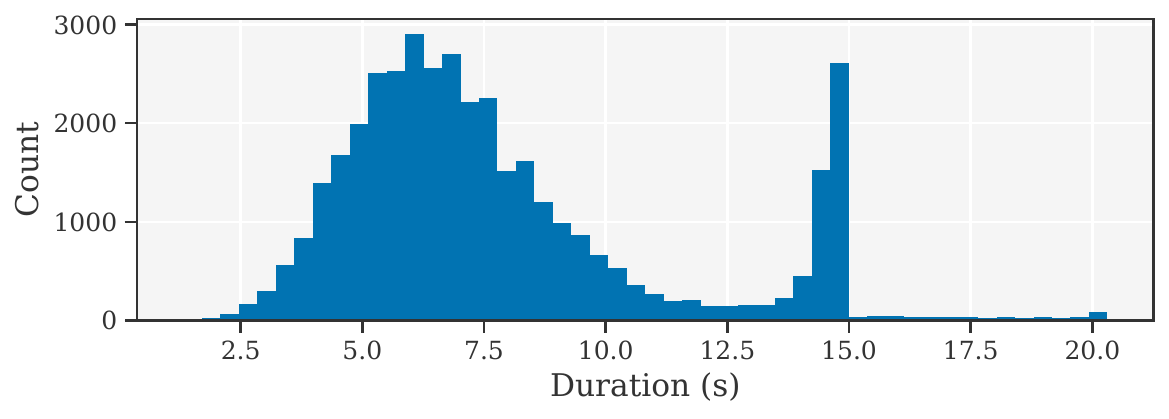}
\vspace{-0.75cm}
\caption{Duration distribution.}
\label{fig:audio_distribution}
\end{figure}

We merge these heterogeneous databases, totaling 85.37 hours of audio. The duration distribution of samples is depicted in Figure~\ref{fig:audio_distribution}, showing that audios are generally shorter than 20 seconds and roughly normally distributed around 6 seconds, with a smaller cluster near 15 seconds from MediaSpeech.

We use the term ``sex'' because existing labels may not reflect gender identity.
We group entries into female, male, or unknown categories. 
The latter denotes samples with no sex label; MediaSpeech provides none, so its Spanish and Turkish subsets are excluded here. Of the remaining 34{,}036 samples, 11{,}128 are female (33\%), 8{,}144 male (24\%), and 14{,}764 (43\%) unknown.

\subsection{Models}

\begin{table*}[h]
\caption{Evaluated models' architecture, number of parameters, language coverage, owner, and license. CU: commercial use permitted.}
\vspace{-0.1cm}
\centering
\resizebox{\textwidth}{!}{%
\begin{tabular}{llccllc}
\hline
\textbf{Model} & \textbf{Architecture / Type} & \textbf{Parameters} & \textbf{Languages} & \textbf{Owner} & \textbf{License} & \textbf{CU} \\
\hline
whisper-large-v3 & Encoder-Decoder & $\sim$1.5B & 99+ & OpenAI & Apache-2.0 & $\checkmark$ \\
Voxtral-Mini-3B-2507 & Speech-LLM & $\sim$3B & 8 & Mistral AI & Apache-2.0 & $\checkmark$ \\
Voxtral-Mini-4B-Realtime-2602 & Causal Speech-LLM & $\sim$4B & 13 & Mistral AI & Apache-2.0 & $\checkmark$ \\
omniASR-CTC-1B-v2 & Wav2Vec2 + CTC Head & $\sim$1B & 1600+ & Meta & Apache-2.0 & $\checkmark$ \\
omniASR-LLM-1B-v2 & Wav2Vec2 + LLM Decoder & $\sim$1B & 1600+ & Meta & Apache-2.0 & $\checkmark$ \\
seamless-m4t-v2-large & w2v-BERT~2.0 + NLLB Decoder & $\sim$2.3B & 101 & Meta & CC BY-NC 4.0 & $\times$ \\
canary-1b-v2 & FastConformer + Decoder & $\sim$978M & 25 & NVIDIA & CC BY 4.0 & $\checkmark$ \\
parakeet-tdt-0.6b-v3 & FastConformer + TDT & $\sim$600M & 25 & NVIDIA & CC BY 4.0 & $\checkmark$ \\
Phi-4-multimodal-instruct & Multimodal-LLM & $\sim$5.6B & 24 & Microsoft & MIT & $\checkmark$ \\
Qwen3-ASR-1.7B & Speech-LLM & $\sim$1.7B & 30+ & Alibaba & Apache-2.0 & $\checkmark$ \\
scribe-v2 & Unknown (API) & - & 90+ & ElevenLabs & Commercial & $\checkmark$ \\
\hline
\end{tabular}
}
\label{tab:asr_models}
\end{table*}

We evaluate ten open-weight models and one commercial API baseline, summarised in Table~\ref{tab:asr_models}. All open-weight models are constrained to fewer than 7B parameters; where a family offers multiple sizes, we select the checkpoint closest in scale to Whisper-large-v3. All but one (seamless-m4t-v2-large, CC BY-NC 4.0) permit commercial use.

\textbf{Whisper-large-v3}~\cite{radford2023robust}, an encoder-decoder Transformer, remains a widely adopted multilingual baseline. Mistral releases two speech-LLMs that share a Ministral~3B backbone~\cite{liu2026ministral3}. \textbf{Voxtral-Mini-3B-2507}~\cite{liu2025voxtral} pairs it with a fine-tuned Whisper-large-v3 encoder and supports audio-conditioned instruction-following. \textbf{Voxtral-Mini-4B-Realtime-2602}~\cite{liu2026voxtral} instead uses a from-scratch causal encoder for native streaming, exposing a configurable latency delay that trades inference time against accuracy. Meta releases three models. The omni pair shares one wav2vec2-style encoder and offers the broadest language coverage in our benchmark. The models differ in decoder: \textbf{omniASR-CTC-1B-v2}~\cite{omnilingual2025omnilingual} uses a CTC head while \textbf{omniASR-LLM-1B-v2}~\cite{omnilingual2025omnilingual} uses a language-model decoder. \textbf{seamless-m4t-v2-large}~\cite{seamless2023} couples a w2v-BERT~2.0 encoder with a fine-tuned NLLB decoder. NVIDIA's \textbf{canary-1b-v2} and \textbf{parakeet-tdt-0.6b-v3}~\cite{sekoyan2025canary} share a FastConformer encoder. The former uses a Transformer decoder and also supports translation. The latter uses a throughput-optimised TDT head that natively emits punctuation, casing, and word-level timestamps. \textbf{Phi-4-multimodal-instruct}~\cite{abouelenin2025phi} routes audio through a conformer encoder and adapter into the per-modality LoRA-tuned Phi-4-Mini LLM. \textbf{Qwen3-ASR-1.7B}~\cite{shi2026qwen3}, built on Qwen3-omni, covers 30 languages and 22 Chinese dialects. \textbf{Scribe~v2}~\cite{elevenlabs_scribe_v2}, a commercial ElevenLabs API, has an undisclosed architecture and training procedure. We include it as a proprietary reference.

\subsection{Evaluation}
We evaluate recognition accuracy with $\text{WER} = (S + D + I)/N$, where $S$, $D$, and $I$ are the numbers of substitutions, deletions, and insertions, and $N$ is the number of reference words. Given the language
imbalance, a corpus-level aggregate would be dominated by the largest split. Thus, we report the \emph{macro-averaged} WER, the unweighted mean of the per-language scores:
\vspace{-0.2cm}
\begin{equation}
    \text{WER}_{\text{macro}} = \frac{1}{L}\sum_{l=1}^{L}\text{WER}_l,
    \vspace{-0.2cm}
\end{equation}
thus, each of the $L$ languages contributes equally regardless sample count. References and hypotheses are normalized by lowercasing and removing punctuation, and digits are converted to words using \texttt{num2words}\footnote{\url{https://pypi.org/project/num2words/}} (not applied to Turkish, which is unsupported). Thus, WER reflects recognition differences rather than formatting mismatches.

%

For efficiency, we report the RTF and its inverse, RTFx:
\begin{equation}
\text{RTF} = \frac{t_{\text{inference}}}{t_{\text{audio}}}, \qquad \text{RTFx} = \frac{1}{\text{RTF}}
\end{equation}
where $t_{\text{inference}}$ is the inference time, and $t_{\text{audio}}$ is the audio duration. RTF below 1.0 (RTFx above 1.0) indicates faster-than-real-time processing; higher RTFx means greater throughput.

All models run on a single NVIDIA GeForce RTX 3090 (24 GB VRAM). For fair and reproducible timing, we process sequentially utterances. For models with configurable delay (e.g., Voxtral-Mini-4B-Realtime), we use the default setting.

\section{Results}

\begin{table*}[h!]
\caption{Overall ASR performance. WER and the substitution, deletion, and insertion rates are macro-averaged across languages and expressed per reference word. The three rates sum to WER and are comparable across systems. Median RTF and RTF\texttimes\ are computed across utterances. Best value per column in bold; Scribe~v2 is shown as a proprietary reference and excluded from the comparison.}
\vspace{-0.15cm}
\centering
\resizebox{\textwidth}{!}{%
\begin{tabular}{lcccccc}
\toprule
\textbf{Model} & \textbf{WER ($\downarrow$, \%)} & \textbf{Median RTF ($\downarrow$)} & \textbf{Median RTF\texttimes ($\uparrow$)} & \textbf{Sub rate ($\downarrow$, \%)} & \textbf{Del rate ($\downarrow$, \%)} & \textbf{Ins rate ($\downarrow$, \%)} \\
\midrule
scribe\_v2                     & 6.45  & 0.1244 & 8.0   & 4.03          & 0.71          & 1.71 \\
\midrule
seamless-m4t-v2-large          & \textbf{8.12}  & 0.0900 & 11.1  & \textbf{5.97} & \textbf{0.88} & \textbf{1.27} \\
omniASR\_LLM\_1B\_v2           & 16.64          & 0.1200 & 8.3   & 8.52          & 6.78          & 1.34 \\
whisper-large-v3               & 19.34          & 0.0872 & 11.5  & 12.72         & 4.38          & 2.24 \\
omniASR\_CTC\_1B\_v2           & 21.46          & 0.0096 & 103.7 & 12.37         & 7.63          & 1.46 \\
Voxtral-Mini-3B-2507           & 23.32          & 0.0801 & 12.5  & 17.22         & 3.52          & 2.59 \\
Qwen3-ASR-1.7B                 & 39.89          & 0.1001 & 10.0  & 30.88         & 4.18          & 4.83 \\
Voxtral-Mini-4B-Realtime-2602  & 48.47          & 0.7258 & 1.4   & 32.04         & 12.75         & 3.67 \\
parakeet-tdt-0.6b-v3           & 51.85          & \textbf{0.0066} & \textbf{151.7} & 32.13 & 15.11 & 4.62 \\
Phi-4-multimodal-instruct      & 56.31          & 0.1599 & 6.3   & 36.28         & 9.92          & 10.10 \\
canary-1b-v2                   & 62.05          & 0.0497 & 20.1  & 26.72         & 26.63         & 8.69 \\
\bottomrule
\end{tabular}
}
\label{tab:overall-performance}
\end{table*}

Table~\ref{tab:overall-performance} reports the overall results. Scribe~v2 attains the lowest WER (6.45\%), with seamless-m4t-v2-large (8.12\%) the best open-weight system, although it cannot be used for commercial purposes. Together, they fare roughly 8 points ahead of the leading freely commercially usable models. Among the latter, omniASR-LLM-1B-v2 (16.64\%), Whisper-large-v3 (19.34\%), omniASR-CTC-1B-v2 (21.46\%), and Voxtral-Mini-3B-2507 (23.32\%) lead. A clear gap separates these from Qwen3-ASR-1.7B (39.89\%) and the remaining systems (WER\,$>$\,48\%), whose errors largely reflect language-coverage gaps. Notably, Phi-4-multimodal-instruct, the largest open-weight model evaluated, reaches only 56.31\% WER at a moderate RTFx of 6.3. It offers no advantage in accuracy, speed, or per-language robustness and underscoring the gap in language coverage.

\subsection{Error Composition and Architectural Patterns}
On the shared Omni-lingual encoder, replacing the CTC head with an LLM decoder lowers all three error rates (subs 12.37$\to$8.52, dels 7.63$\to$6.78, ins 1.46$\to$1.34), so the decoder change helps uniformly rather than trading one error type for another. Across the full set, insertions and deletions rise together for the weakest, low-coverage systems, with Phi-4-multimodal-instruct (ins 10.10) and canary-1b-v2 (ins 8.69, dels 26.63) at the extreme, while the accurate tier keeps insertions near 1.3--2.6 regardless of decoder type. Substitutions dominate the error budget of every functioning system.

\subsection{Speed and WER Trade-offs}
\begin{figure}[h]
    \centering
    \includegraphics[width=1.0\linewidth]{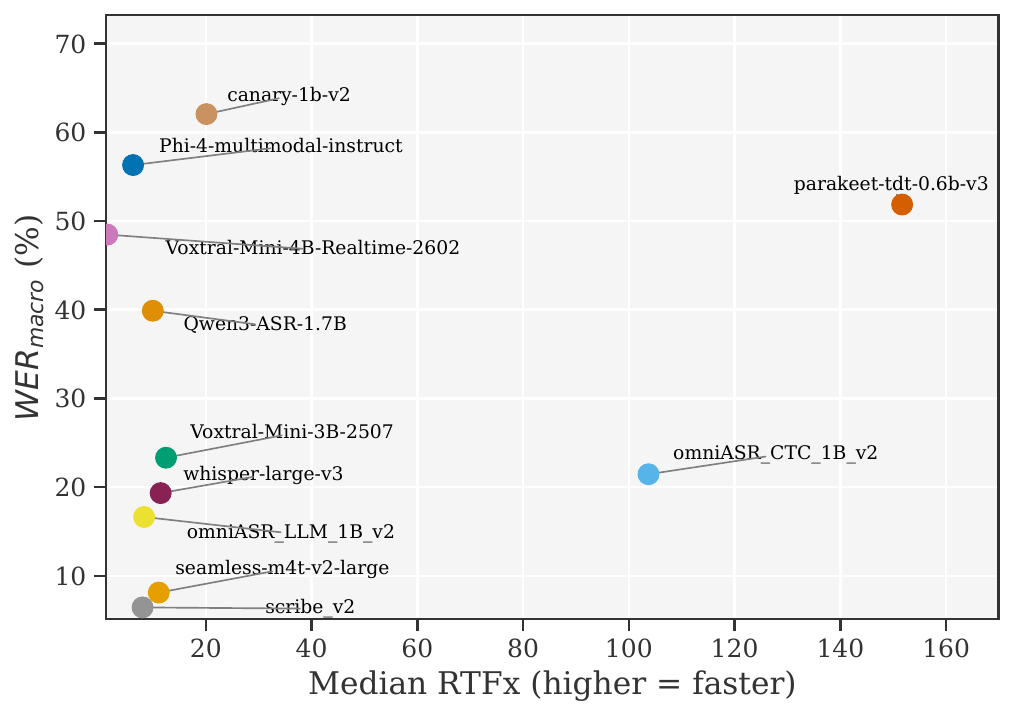}
    \vspace{-0.7cm}
    \caption{Performance ($\mathrm{WER}_{\mathrm{macro}}$) vs.\ $\mathrm{Efficiency}$ (RTFx).}
    \label{fig:wer_vs_rtf}
\end{figure}

Figure~\ref{fig:wer_vs_rtf} shows the speed-accuracy trade-off. parakeet-tdt-0.6b-v3 attains the highest throughput (RTFx~151.7), followed by omniASR\_CTC\_1B\_v2 (RTFx~103.7), both notably exceeding real time. However, parakeet's WER of 51.9\%, a consequence of its limited language coverage, restricts its multilingual applicability. omniASR\_CTC\_1B\_v2 thus offers the best accuracy--efficiency trade-off among freely usable models, pairing competitive WER (21.5\%) with exceptional throughput. At the opposite extreme, Voxtral-Mini-4B-Realtime-2602 posts the lowest RTFx~(1.4) and a high macro WER~(48.5\%); although designed for streaming with configurable latency, its causal encoder likely incurs overhead in offline batch evaluation.

\subsection{Per-Language and Per-Sex Analysis}
\begin{figure*}[h]
    \vspace{-0.1cm}
    \centering
    \includegraphics[width=1\textwidth]{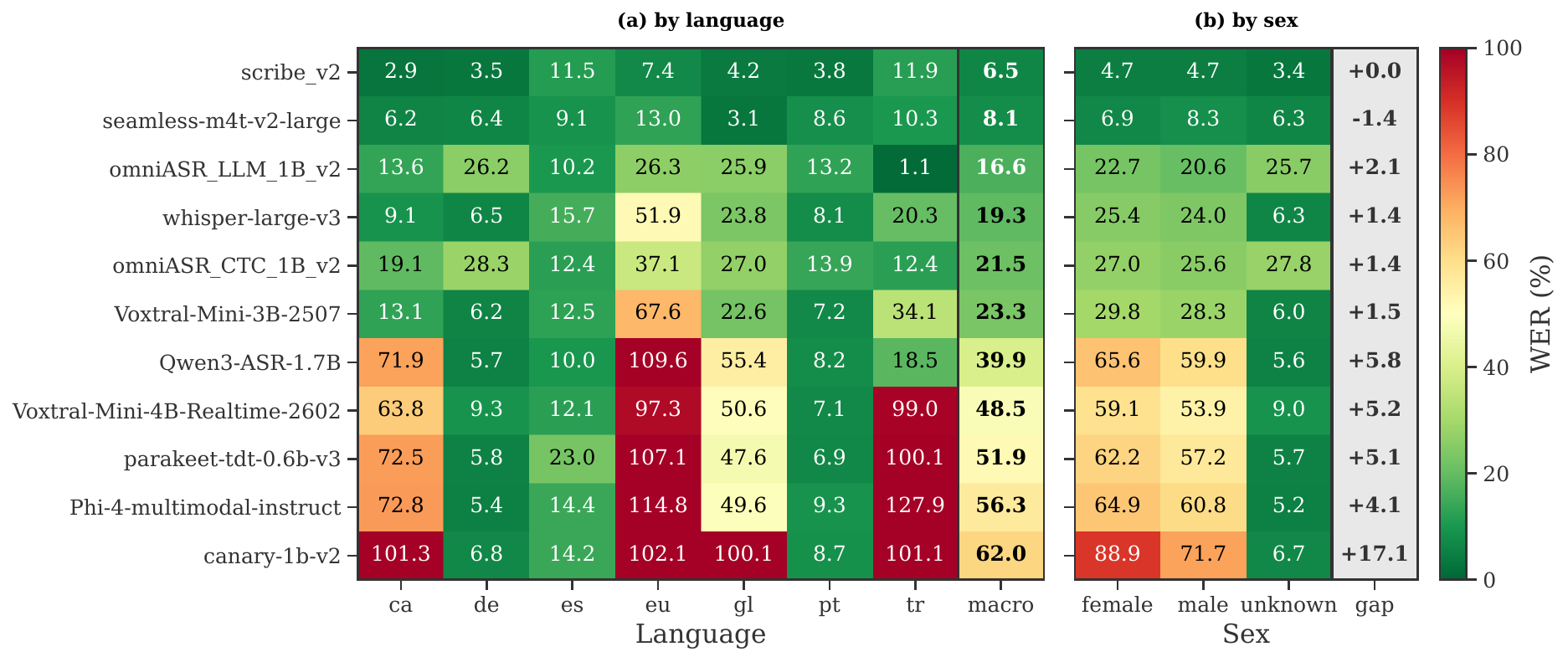}
    \vspace{-0.75cm}
    \caption{Model's WER: (a) across languages and (b) per speaker sex. Models are ordered by macro WER.}
    \label{fig:heatmap}
\end{figure*}

Figure~\ref{fig:heatmap}(a) reveals two regimes. For high-resource languages (German, Spanish, Portuguese), most systems cluster in a narrow WER band, limiting discriminative power. Comparisons here should be read with caution, as the datasets differ in domain ( see Table~\ref{tab:datasets}). Such mismatches explain counter-intuitive orderings, e.g.\ higher Spanish than German WER despite Spanish being well represented.

For lower-resource languages, performance diverges sharply.
%
%
Basque is consistently the most challenging language. While this likely reflects limited representation in pretraining corpora, other factors may also contribute, including morphology, tokenization mismatches or acoustic domain.
Systems without explicit Basque coverage (parakeet-tdt-0.6b-v3, canary-1b-v2) degrade to near-random output. Catalan is also consistently difficult, with elevated substitution rates pointing to under-representation. Galician and Turkish are intermediate: top-tier models cope well while weaker ones show high variance, e.g.\ Phi-4-multimodal-instruct records the single worst score on Turkish (127.9\%). The omniASR family is the most consistent cross-lingually, including on Basque, a direct effect of its training coverage. Scribe~v2 and seamless-m4t-v2-large stay stable across languages, and their gap over the open-weight tier is widest on the lower-resource set.

Figure~\ref{fig:heatmap}(b) shows a consistent male advantage across most models, largest for canary-1b-v2 (gap $+17.1$); seamless-m4t-v2-large is the exception, slightly favouring female speakers. 
%
Scribe~v2 exhibits minimal observed differences between male and female speakers, though its opaque training data and the dataset imbalances make this parity hard to attribute.
%
%
The \textit{unknown} group consists of German only, since MediaSpeech provides no sex labels and its Spanish and Turkish subsets are excluded from this analysis. Thus, lower WER in this category reflects dataset composition rather than intrinsic robustness.

\vspace{-0.2cm}
\section{Limitations}
\vspace{-0.2cm}
Several limitations apply. First, the datasets are heterogeneous in domain and recording conditions, so cross-language comparisons should be interpreted with caution despite transcription normalization; broadcast-derived sets likely inflate WER relative to read speech. Second, we cannot confirm whether models were trained on benchmark subsets, which would optimistically bias their WER. Third, inference times were measured on identical hardware without architecture-specific configuration, possibly disadvantaging some architectures; for Scribe~v2, RTF and RTFx instead reflect API latency. Fourth, language labels can mask dialectal variation (e.g., Brazilian vs.\ European Portuguese). A controlled assessment of demographic disparities would require datasets balanced across speaker groups and languages, beyond this benchmark's scope. Nonetheless, the consistency of these disparities across architectures suggests fairness challenges are pervasive in multilingual ASR.

\section{Conclusions}
We benchmarked ten open-weight and one commercial ASR system on Iberian languages, with Turkish and German as controls, assessing accuracy, efficiency, and error composition. The resulting hierarchy is conditioned by licensing. Scribe~v2 achieves the best $\mathrm{WER}_{\mathrm{macro}}$, and seamless-m4t-v2-large is the strongest open-weight system but is restricted to non-commercial use. Among freely usable models, omniASR\_LLM\_1B\_v2 and Whisper-large-v3 lead on accuracy, with the former also the most robust across low-resource and typologically diverse languages. omniASR\_CTC\_1B\_v2 offers the best accuracy--efficiency trade-off (RTFx~103), whereas parakeet-tdt-0.6b-v3 attains the highest throughput (RTFx~152) at substantially degraded WER. Per-language results establish language coverage as a first-order selection criterion: high-resource languages (German, Spanish, Portuguese) yield consistent performance, whereas lower-resource ones, Basque in particular, expose sharp divergences, with uncovered models degrading severely. Speaker-sex disaggregation further reveals a consistent male advantage, with only Scribe~v2 and seamless-m4t-v2-large remaining equitable. Overall, no single open-weight model dominates: users must navigate trade-offs among accuracy, speed, language coverage, and licensing, a clear opportunity for the open-weight ASR community.

\pagebreak

\section{Generative AI Use Disclosure}
We used a generative AI to assist in paraphrasing and improving clarity and grammar in parts of the manuscript; all generated content was carefully reviewed and validated by the authors.



\end{document}